\documentclass{article}       
\usepackage{natbib}
\bibpunct{[}{]}{,}{n}{}{;}

\usepackage{footmisc}

\usepackage{graphicx}
\usepackage{mathtools,caption}
\usepackage{amssymb}
\usepackage{eucal}
\usepackage{amsmath}
\usepackage{graphicx}
\usepackage{float}

\newcommand{\wt}{\widetilde}
 \numberwithin{equation}{section} 
  
\begin{document}

\title{A New Training Method for Feedforward Neural Networks Based on
Geometric Contraction Property of Activation Functions}
\author{Petre Birtea\footnote{Department of Mathematics, West University of Timi\c soara; Bd. V. P\^ arvan, No 4, 300223 Timi\c soara, Rom\^ ania; Email: petre.birtea@e-uvt.ro}, Cosmin Cern\u azanu-Gl\u avan\footnote{Department of Computer Science, "Politehnica" University of Timi\c soara; Bd. V. P\^ arvan, No 2, 300223 Timi\c soara, Rom\^ ania; Email: cosmin.cernazanu@cs.upt.ro; sisu.eugen@gmail.com}, Alexandru \c Si\c su\footnotemark[2]
}
%
%
%

\date{}

\maketitle

\begin{abstract}
We propose a new training method for a feedforward neural network having the activation functions with the geometric contraction property. The method consists of constructing a new functional that is less nonlinear in comparison with the classical functional by removing the nonlinearity of the activation function from the output layer. We validate this new method by a series of experiments that show an improved learning speed and better classification error.
\end{abstract}
\section{Introduction}
From the mathematical point of view a feedforward neural network (FNN) is a sequence of mathematical operations (of type vector multiplication and non-linear activation) able to transform vectors from input space into vectors from output space.

A feed forward neural network with $n$ hidden layers can be seen as a minimization problem of the following cost function:
\begin{equation}\label{functionala_principala}
E({\bf W}_1,{\bf W}_2,...,{\bf W}_n):=\sum_{\alpha=1}^{T}||{\bf s}_n({\bf W}_n\cdot...\cdot{\bf s}_2({\bf W}_2\cdot{\bf s}_1({\bf W}_1\cdot{\bf x}_\alpha))...)-{\bf y}_\alpha||^{2},
\end{equation}
where:

$\bullet$ ${\bf W}_i$ represents the weights matrices between the $(i-1)^{th}$ layer and $i^{th}$ layer.

$\bullet$ the vectorial activation function ${\bf s}_i:\mathbb{R}^{d_i}\rightarrow\mathbb{R}^{d_i}$ is defined by, ${\bf s}_i({\bf x}) := (s(x_1) , ... , s(x_{d_i}))$, where $d_i$ is the number of neurons on the $i^{th}$ layer, and $s:\mathbb{R}\rightarrow\mathbb{R}$ is a neuron activation function, and ${\bf x}=(x_1,x_2,...,x_{d_i})$ is the numeric codified neural network input \par

$\bullet$ ${\bf x}_\alpha$ is the $\alpha^{th}$ input vector of the training set

$\bullet$ ${\bf y}_\alpha$ is the $\alpha^{th}$ output vector of the training set

$\bullet$ $T$ is the number of samples in the training set

The training of an neural network means to find the weights matrices ${\bf W}_1,{\bf W}_2,...,{\bf W}_n$ that render the minimum value of the cost function $E$. The most common approach is to apply an iterative numerical algorithm that in the end will generate a set of approximated optimal values $({\bf {W}}_1^{(opt)},..,{\bf {W}}_n^{(opt)})$.

Usually the activation function $s$ is a non-linear function that increases the amount and the complexity of the computations necessary to solve the above minimization problem.

This cost function can be seen as a {\bf distance function in an Euclidean space of the weights variables}.
Hence the minimization of cost function can be translated into the problem of minimization  the distance between the target values and the output of the neural network. 

\begin{figure}[H]
	\centering
	\includegraphics[width=0.7\textwidth]{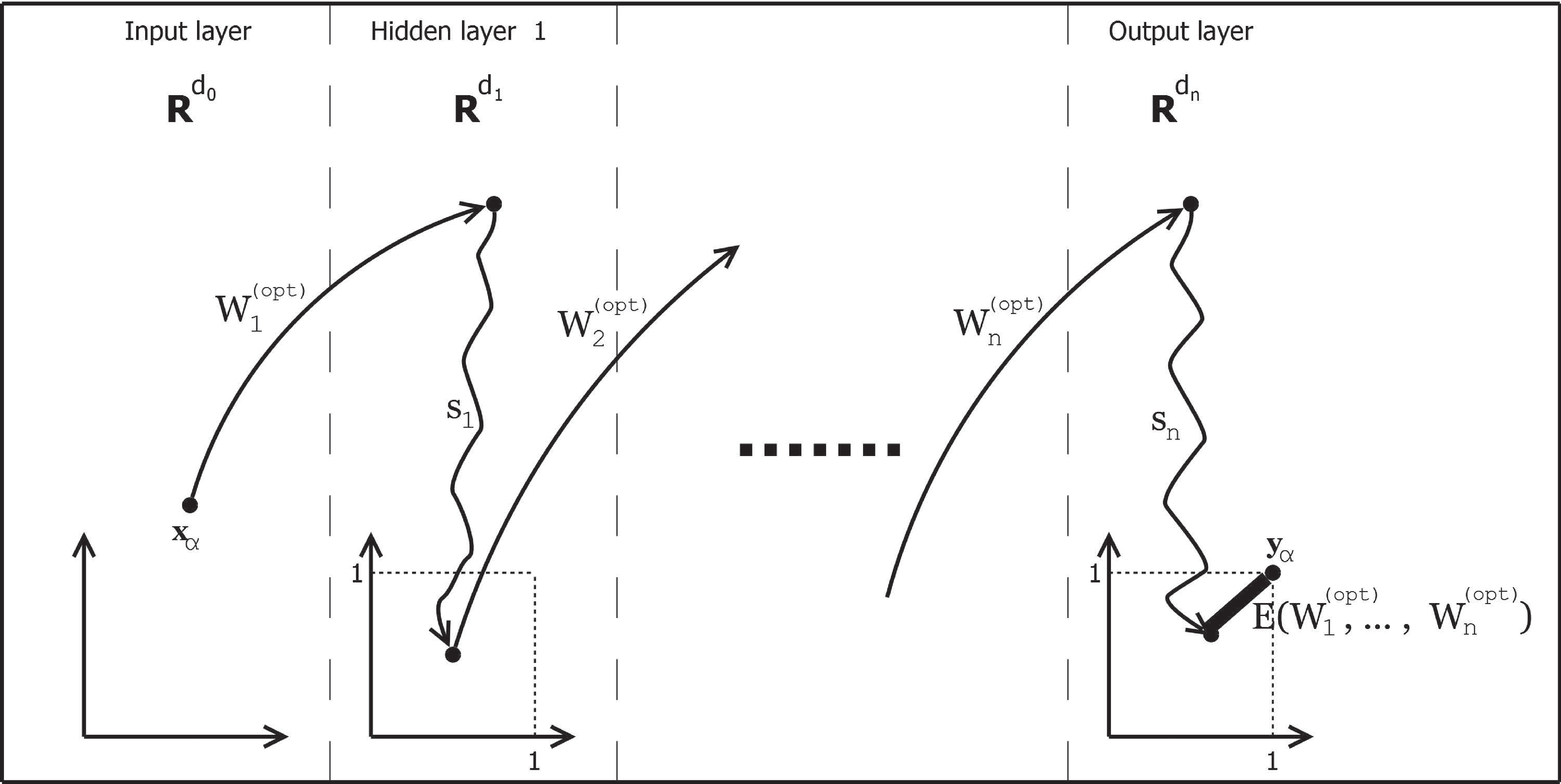}
	\caption{An intuitive geometric depiction of the activation function and the error distance in a classical training scenario.}
	\label{fig:drawing_classical}
\end{figure}

Figure \ref{fig:drawing_classical} presents an intuitive depiction of the linear transformation followed by the non-linear transformation (the activation function) occurring between two consecutive layers. The cost function can be seen at the end of the transformation process.
Since the development of the backpropagation algorithm (BP) there was a lot of effort in order to address the shortcoming of applying the BP in really large networks. The shortcomings were vanishing gradient problem and the algorithm complexity translated into CPU time. This problems were tackled by  improvements that followed several directions, directions that we might classify into : mathematical, structural and algorithmic. 

When talking about  mathematical improvements we have to mention: Quasi Newton methods \cite{Broyden65}, \cite{Fletcher63}, \cite{Goldfarb70}, \cite{Shannon70}; Conjugate gradient descent \cite{Hestenes52}, \cite{Moller93}. However this methods turned out to be expensive from a computational point of view. Methods like Hessian-free optimization \cite{Moller93}, \cite{Pearlmutter94}, \cite{Schraudolph2002}, \cite{Martens2010} address the problem of vanishing gradient in feed forward neural networks. Levenberg-Marquardt \cite{Levenberg44}, \cite{Marquardt63}, \cite{Schaback92} improves the convergence speed.

Another set of improvements came from the modification of the structure of a FNN like in the case of the Dropout method \cite{Hinton2012b}, \cite{Ba2013}, \cite{Baldi2014} or by modifying the activation function from non-linear to linear like in the case of Rectified Linear Units (ReLU) \cite{Malik90}, \cite{Nair2010}, \cite{Maas2013}, \cite{Glorot2011}, \cite{Krizhevsky2012}, \cite{Dahl2013}.

Numeric and algorithmic optimisation techniques were also employed when trying to optimize the whole process of running BP: regularization and weight decay \cite{Hanson89}, \cite{Weigend91}, \cite{Krogh92}, momentum \cite{Rumelhart86}, \cite{Fahlman88}, \cite{West95}, Nesterov accelerated gradient descent \cite{Nesterov83}. Other gradient based optimization methods turned out to be really efficient: RmsProp \cite{Tieleman2012} \cite{Schaul2013}, Adagrad \cite{Duchi2011}, Adadelta \cite{Zeiler2012}.
\section{Canceling the non linearity on the output layer}
Our idea is to modify the initial cost function by taking away the non-linearity of the activation function of  the output layer. That leads to reduce the amount of computation for finding the optimal values of the weights matrices that renders a minimum value of the cost function. In order to do this, we need the  following hypothesis (for example, the sigmoid function verifies these hypothesis):

$\bullet$ the activation function $s$ is an diffeomorphism on its domain of definition and moreover it is a contraction (i.e. the derivative of $s$ is strictly smaller than 1 on the whole interval of definition).

$\bullet$ the coordinates of output vectors $\wt{\bf y}_\alpha$ belong to the co-domain of activation function $s$. The output vectors  $\wt{\bf y}_\alpha$ can be the same with the output vectors ${\bf y}_\alpha$ if the activation function $s$ allows it, see the discussion in next section. 

More precisely, we propose the following modified cost function:

\begin{equation}\label{functionala_modificata}
\wt{C}({\bf \wt{W}}_1,{\bf \wt{W}}_2,...,{\bf \wt{W}}_n):=\sum_{1}^{T}||{\bf \wt{W}}_n\cdot{\bf s}_{n-1}({\bf \wt{W}}_{n-1}\cdot...\cdot{\bf s}_2({\bf \wt{W}}_2\cdot{\bf s}_1({\bf \wt{W}}_1\cdot{\bf x}_\alpha))...)-{\bf s}_n^{-1}(\wt{\bf y}_\alpha)||^{2},
\end{equation}
where:

$\bullet$ ${\bf s}_n^{-1}:\mathbb{R}^{d_n}\rightarrow\mathbb{R}^{d_n}$ is the inverse of the vectorial activation function ${\bf s}_n$

$\bullet$ ${\bf \wt{W}}_i$ represents the weights matrices between the $(i-1)^{th}$ layer and $i^{th}$ layer.

$\bullet$ ${\wt{\bf y}}_\alpha$ is the $\alpha^{th}$ output vector of the training set

We propose to train the neural network in the same manner, but using the newly introduced cost function $\wt{C}$. At the end will obtain a set of approximated optimal values $({\bf \wt{W}}_1^{(opt)},..,{\bf \wt{W}}_n^{(opt)})$ for the variables that we are interested: in our case the weights matrices.

\begin{figure}[H]
	\centering
	\includegraphics[width=0.7\textwidth]{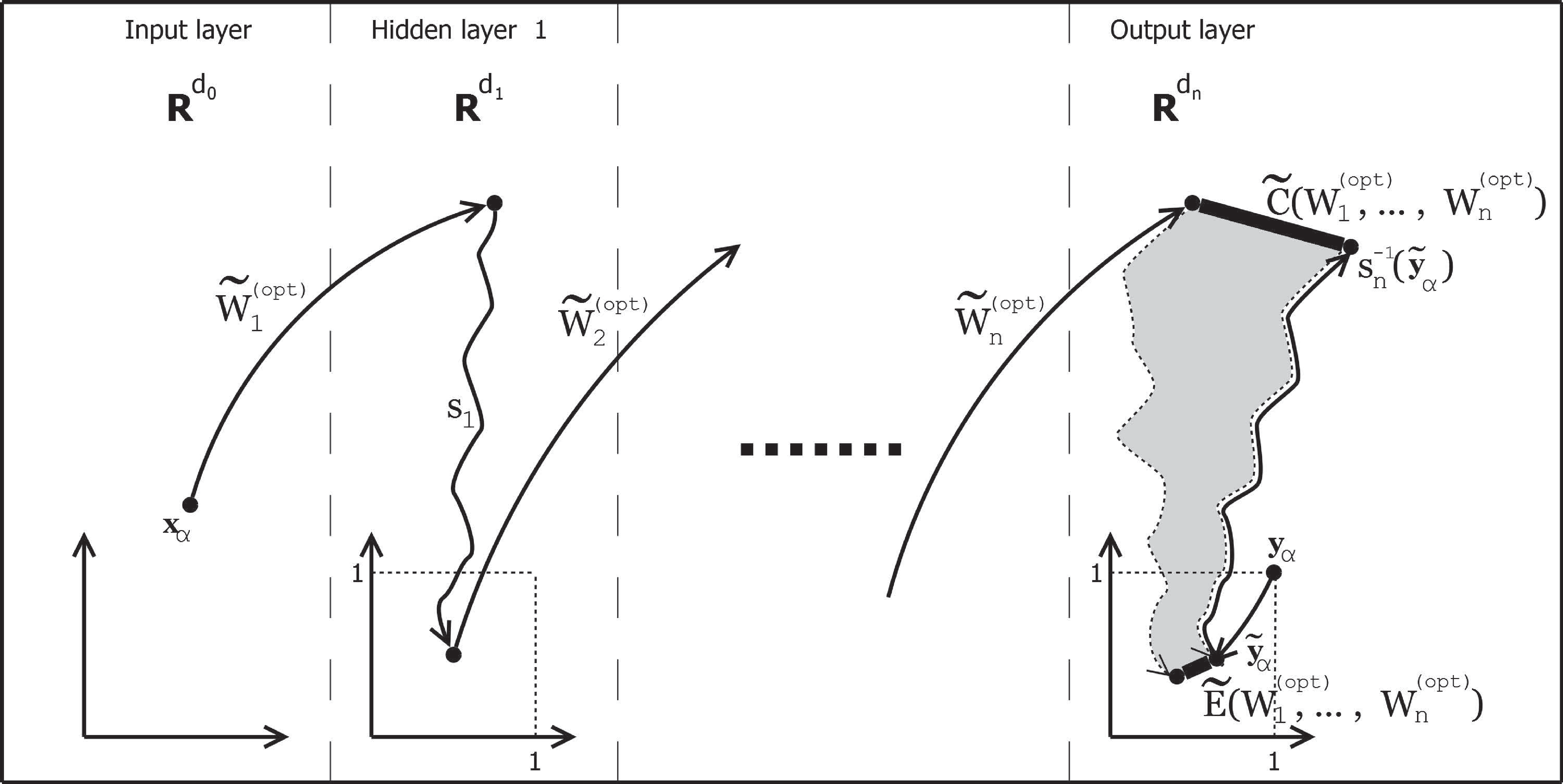}
	\caption{An intuitive geometric depiction of the activation function and the error distance in the newly proposed method.}
	\label{fig:drawing_trick}
\end{figure}

We will show that introducing the computed weights $({\bf \wt{W}}_1^{(opt)},..,{\bf \wt{W}}_n^{(opt)})$ in the typical feedforward error function \eqref{functionala_principala}  $$\wt{E}({\bf \wt{W}}_1,{\bf \wt{W}}_2,...,{\bf \wt{W}}_n):=\sum_{1}^{T}||{\bf s}_n({\bf \wt{W}}_n\cdot...\cdot{\bf s}_2({\bf \wt{W}}_2\cdot{\bf s}_1({\bf \wt{W}}_1\cdot{\bf x}_\alpha))...)-\wt{\bf y}_\alpha||^{2},$$ where the new output vectors are $\wt{\bf y}_\alpha$, will yield even better values. Indeed, due to the fact that vectorial function $\bf{s}_n$ is a contraction this implies that it will contract the distances, see Figure \ref{fig:drawing_trick}. Expressed in a mathematical form we have: 
$$\wt{C}({\bf \wt{W}}_1^{(opt)},..,{\bf \wt{W}}_n^{(opt)}) > \wt{E}({\bf \wt{W}}_1^{(opt)},..,{\bf \wt{W}}_n^{(opt)}) $$
Starting from an initial set of conditions and applying the standard training procedure (using the classical cost function $E$) we obtain a set of weight values: $({\bf {W}}_1^{(opt)},..,{\bf {W}}_n^{(opt)})$, which in general are different than the weights $({\bf \wt{W}}_1^{(opt)},..,{\bf \wt{W}}_n^{(opt)})$ obtained by training the network using the modified cost function $\wt{C}$.

A direct comparison between the two training methods, the classical one using the cost function $E$ and the proposed one using the cost function $\wt{C}$ cannot be performed, as in general the inequality 

\begin{equation}\label{comparatie}
E({\bf {W}}_1^{(opt)},..,{\bf {W}}_n^{(opt)}) > \wt{E}({\bf \wt{W}}_1^{(opt)},..,{\bf \wt{W}}_n^{(opt)}),
\end{equation}

\noindent can not by proved mathematically.

Experimentally, where we take as a criteria of performance the number of rightfully classified examples (accuracy measure), we prove that our proposed training method yields better classification results than the classical one. Also, the experiments show that as a byproduct of this method we obtain also a faster classification for the chosen datasets.

\begin{figure}[H]
	\centering
	\includegraphics[width=0.5\textwidth]{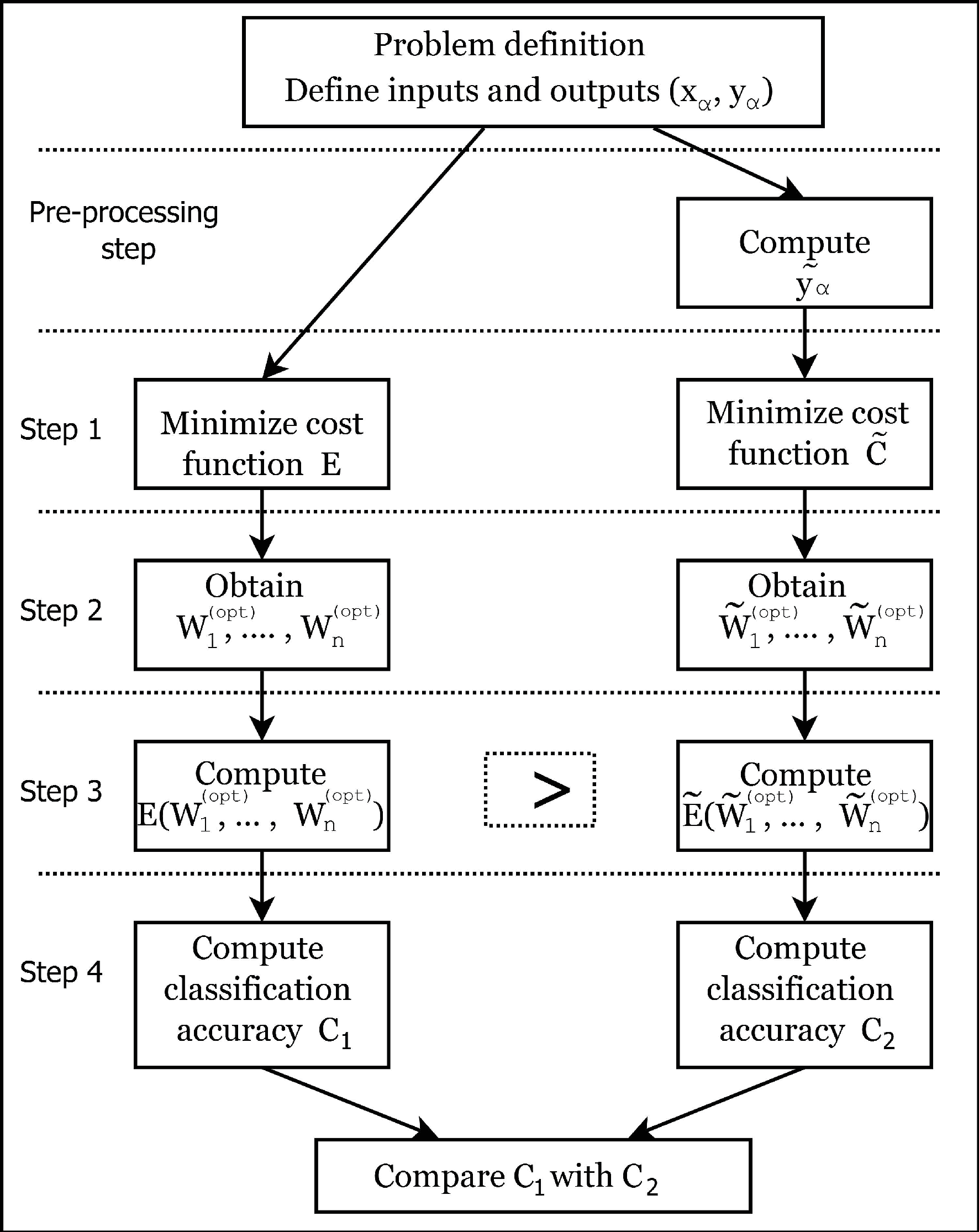}
	\caption{The logical diagram used to compare the classical training method with the newly proposed one.}
	\label{fig:logical_diagram}
\end{figure}

The comparison and the implementation procedure of the classical training method and the newly proposed method follows the logical diagram depicted in Figure \ref{fig:logical_diagram}. Thus, we experimentally compare $C_1$ (the accuracy percentage for the classical method) with $C_2$ (the accuracy percentage for the newly proposed method). The newly proposed method needs an supplementary preprocessing step where we need to define the new output vectors ${\wt{\bf y}}_\alpha$.This intermediate step is necessary only when the coordinates of the output vectors $\bf{y}_\alpha$ in the classical method does not belong to the co-domain of the activation function $s$. The rest of the steps remain the same with the difference that we minimize different cost function $\wt{C}$, defined in \eqref{functionala_modificata}.

Experimentally, one can observed that after {\it Step 3} we have a faster learning (lower value of the respectively cost functions) using the new method. This advantage is reflected at the end of the {\it Step 4}, where one compares the actual accuracy ($C_1$ and $C_2$) obtained by the two methods.

\section{Experiments}
We have done four experiments, in the first three we have used the MNIST dataset \cite{mnist}. This is the most common dataset (actually is a subset of a larger available set - NIST) used in this field and contains a large set of images representing handwrite digits. The training set contains 60.000 examples and the testing set 10.000 examples. Each experiment consists of training a neural network using two different cost functions, the classical cost function $E$ and respectively, the new proposed cost function $\wt{C}$. In the first three experiments we have chosen neural network configurations already used in literature, see \cite{lecun98}. For the forth experiment we have used the notMNIST dataset \cite{notMNIST}. The training set contains 210.000 examples and the testing set 10.000 examples. During all experiments we have obtained a faster learning and also a better classification error using the new proposed cost function.

For the first experiment we consider a feedforward neural network having the following configuration: the input layer contains 784 neurons, followed by two hidden layers with 300, respectively 100 neurons, and an output layer containing 10 neurons (one neuron for each digit).

For the classical training of the first experiment all neurons are classical neurons having as activation function the sigmoid, $s:\mathbb{R}\rightarrow(0,1)$ , $s(x) = (1+e^{-x})^{-1}$. Each target is a vector $\bf {y}_\alpha \in \mathbb{R}^{10}$ with components $0$ or $1$.

For the second training of the first experiment, we start by verifying that the sigmoid function obey the first hypothesis discussed in the previous section, meaning it is a contracting diffeomorphism on its domain of definition.

In order to implement our method, the second hypothesis needs to be verified. We need to apply the inverse sigmoid on each component of the target vector $\bf {y}_\alpha$. As $"s^{-1}(1) = \infty"$ and $"s^{-1}(0) = -\infty"$ one must replace the coordinate entries $0$ and $1$ in $\bf {y}_\alpha$ with other two reference values belonging to $(0,1)$. To find a set of suitable values, a series of experiments were conducted and we obtain the best results (from the training point of view) when replacing $0$ with $0.2227$ and $1$ with $0.7773$. Thus, the coordinate entries of the new output vectors $\bf \wt{y}_\alpha$ are $0.2227$ and $0.7773$. Following the logical diagram depicted in Figure \ref{fig:logical_diagram}, we implement our method accordingly: 

$\bullet$ {\it Pre-processing step}: replace the vector components $0$ and $1$ in the target vector $\bf {y}_\alpha$ with $0.2227$ , respectively $0.7773$, thus obtaining the new output vectors $\bf \wt{y}_\alpha$. Compute the vectors ${\bf s}_n^{-1}({\bf \wt{y}}_\alpha)$ that are used for defining cost function $\wt {C}$.

$\bullet$ {\it Step 1}: apply the gradient descent algorithm for the new cost function $\wt {C}$. 

$\bullet$ {\it Step 2}: after a sufficient number of iterations we obtain the weights matrices ${\bf \wt{W}}_1^{(opt)},{\bf \wt{W}}_2^{(opt)},{\bf \wt{W}}_3^{(opt)}$, where ${\bf \wt{W}}_1^{(opt)}$ is the weights matrix of $784\times 300$ elements, between the input layer and the first hidden layer, and ${\bf \wt{W}}_2^{(opt)}$ is the weights matrix of $300\times 100$ elements, between the first hidden layer and the second hidden layer and ${\bf \wt{W}}_3^{(opt)}$ is the weights matrix of $300\times 10$ elements, between the second hidden layer and the output layer.

\begin{figure}[H]
	\centering
	\includegraphics[width=1.0\textwidth]{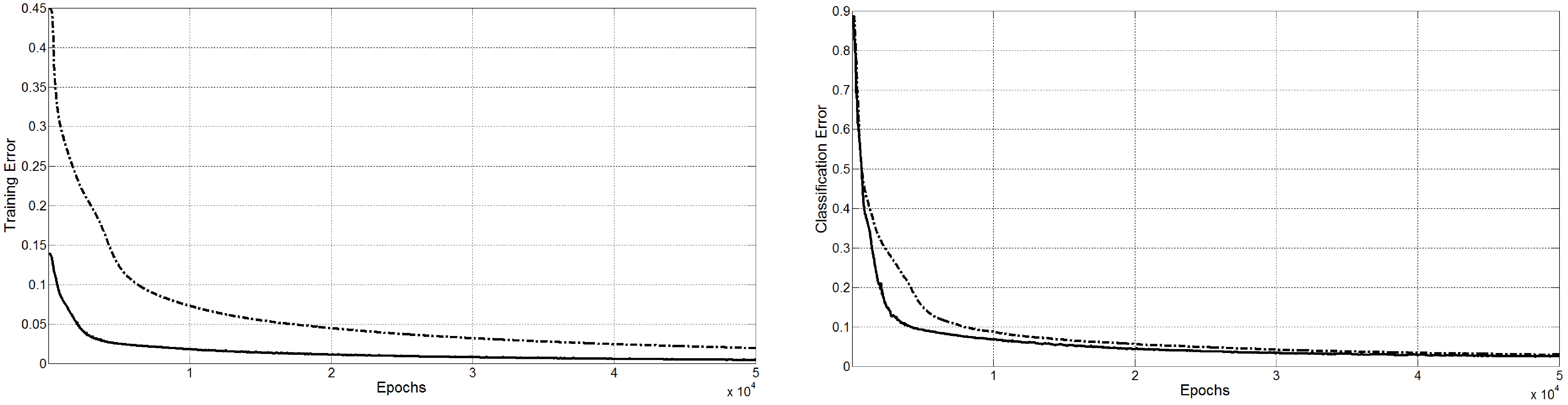}
	\caption{(Left) Training error for experiment 1. The new proposed method (continuous line) shows a smaller error than the classical method (dash line) after $5\times 10^4$ epochs. Also, the new method yields a faster convergence in terms of training error. (Right) Classification error for experiment 1. The new proposed method (continuous line) shows a better classification than the classical method (dash line). After  $5\times 10^4$ epochs, the classification error obtained with the new method is 2.55 vs. 3.02 which is the classification error obtained using the classical method.}
	\label{fig:training_e1}
\end{figure}

$\bullet$ {\it Step 3}: compute the value $\wt{E}({\bf \wt{W}}_1^{(opt)},{\bf \wt{W}}_2^{(opt)},{\bf \wt{W}}_3^{(opt)})$ and compare with $E({\bf {W}}_1^{(opt)},..,{\bf {W}}_n^{(opt)})$ obtained using the classical training, see Figure \ref{fig:training_e1} (Left).

$\bullet$ {\it Step 4}: compute the accuracy value $C_2$ obtained with the new proposed method and compare with accuracy value $C_1$ obtained using the classical training, see Figure \ref{fig:training_e1} (Right). For this configuration, LeCun et al. \cite{lecun98} report a classification error of 3.05, value which we have obtained after 47.750 epochs with classical training. After the same number of epochs, using the new proposed method, we have obtain a classification error of 2.59 .

\begin{figure}[H]
	\centering
	\includegraphics[width=1.0\textwidth]{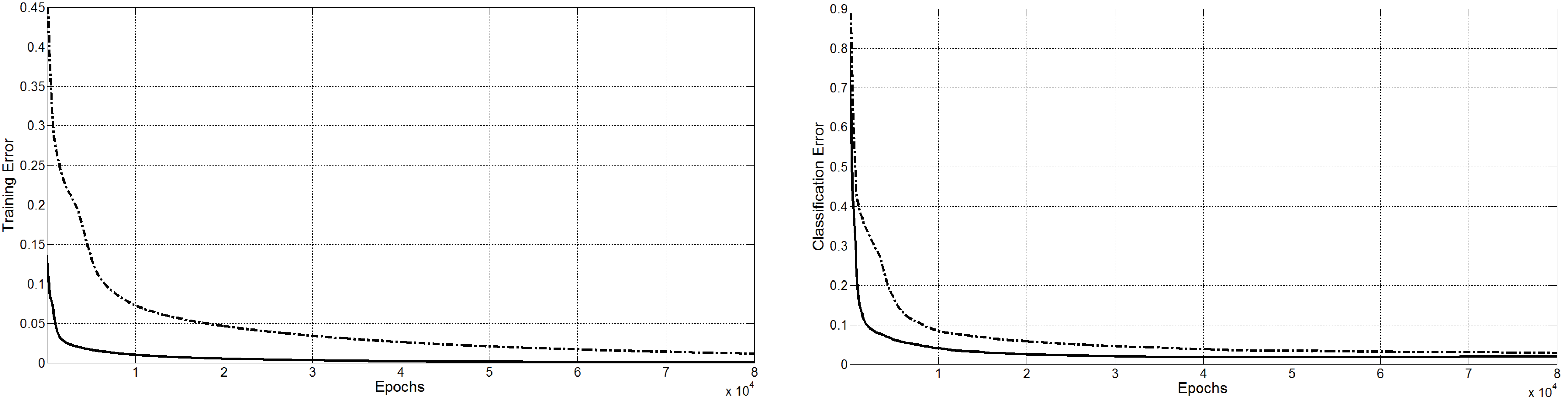}
	\caption{(Left) Training error for experiment 2. The new proposed method (continuous line) shows a smaller error than the classical method (dash line) after $8\times 10^4$ epochs. Also, the new method yields a faster convergence in terms of training error. (Right) Classification error for experiment 2. The new proposed method (continuous line) shows a better classification than the classical method (dash line). After  $8\times 10^4$ epochs, the classification error obtained with the new method is 1.99 vs. 2.88 which is the classification error obtained using the classical method.}
	\label{fig:training_e2}
\end{figure}

For the second experiment we changed the architecture of the neural network in the following way: the input layer contains 784 neurons, followed by two hidden layers with 500, respectively 150 neurons, and an output layer containing 10 neurons (one neuron for each digit). Both classical and the new proposed training followed the same procedure as in the first experiment were we have described them in details. The comparison between the two trainings of the second experiment can be seen in Figure \ref{fig:training_e2}. For this configuration, LeCun et al. \cite{lecun98} report a classification error of 2.95, value which we have obtained after 77.800 epochs with classical training. After the same number of epochs, using the new proposed method, we have obtain a classification error of 2.00 .

\begin{figure}[H]
	\centering
	\includegraphics[width=1.0\textwidth]{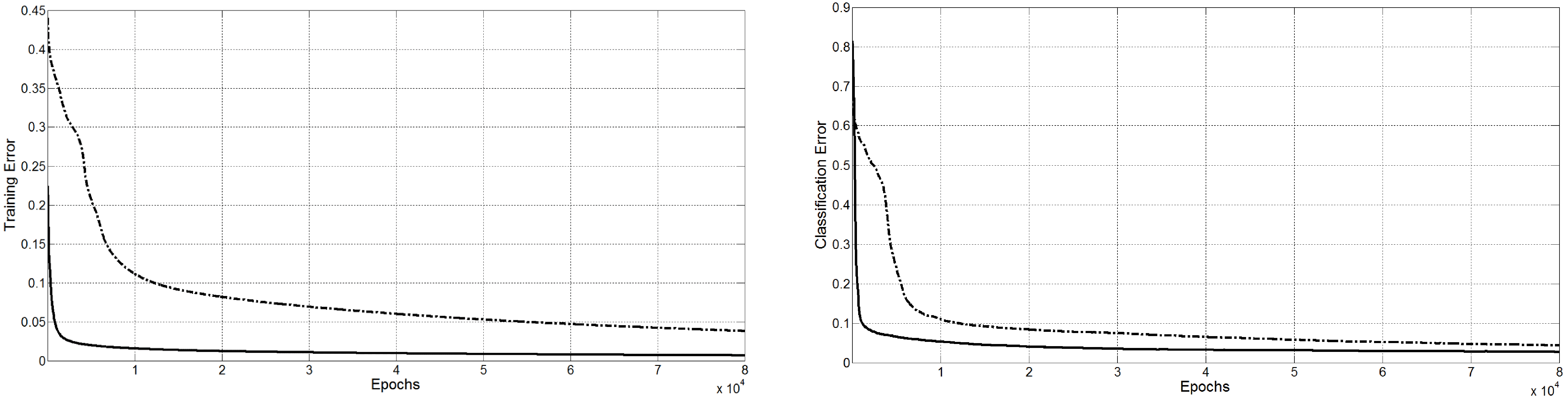}
	\caption{(Left) Training error for experiment 3. The new proposed method (continuous line) shows a smaller error than the classical method (dash line) after $8\times 10^4$ epochs. Also, the new method yields a faster convergence in terms of training error. (Right) Classification error for experiment 3. The new proposed method (continuous line) shows a better classification than the classical method (dash line). After  $8\times 10^4$ epochs, the classification error obtained with the new method is 2.76 vs. 4.39 which is the classification error obtained using the classical method.}
	\label{fig:training_e3}
\end{figure}

In the third experiment we considered an architecture of the neural network with only a single hidden layer of 1000 neurons. Thus, the architecture is as follows: the input layer contains 784 neurons, followed by one hidden layer with 1000 neurons, and an output layer containing 10 neurons (one neuron for each digit). Both classical and the new proposed training followed the same procedure as in the first two experiments. The comparison between the two trainings of the third experiment can be seen in Figure \ref{fig:training_e3}. For this configuration, LeCun et al. \cite{lecun98} report a classification error of 4.5, value which we have obtained after 77.050 epochs with classical training. After the same number of epochs, using the new proposed method, we have obtain a classification error of 2.79.

\begin{figure}[H]
	\centering
	\includegraphics[width=1.0\textwidth]{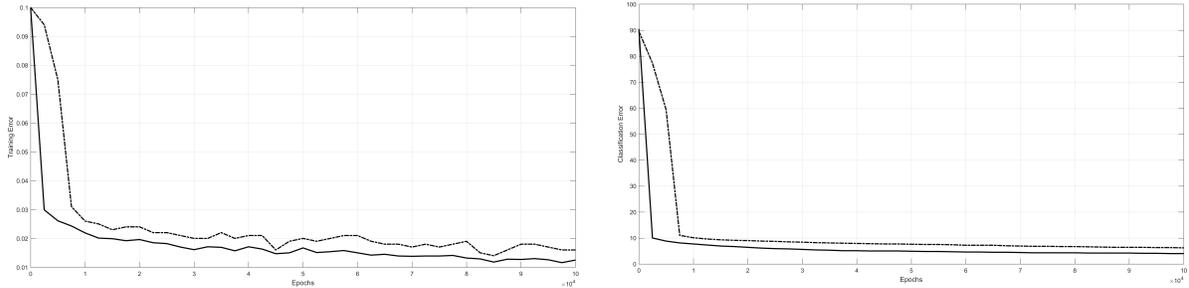}
	\caption{(Left) Training error for experiment 4. The new proposed method (continuous line) shows a smaller error than the classical method (dash line) after $10^5$ epochs. Also, the new method yields a faster convergence in terms of training error. (Right) Classification error for experiment 4. The new proposed method (continuous line) shows a better classification than the classical method (dash line). After  $10^5$ epochs, the classification error obtained with the new method is 4 vs. 6.2 which is the classification error obtained using the classical method.}
	\label{fig:training_e4}
\end{figure}

You have made a forth experiment where we have changed the MNIST dataset with a more challenging one dataset, namely notMNIST \cite{notMNIST}. For this experiment, the architecture of the neural network is the following: input layer contains 784 neurons (images of 28x28 pixels), followed by three hidden layers of 4096, 2048 and respectively 1024 neurons and an output layer of 10 neurons (letters from 'A' to 'J'). As before, we have the same procedure for training classical and the new proposed training method as in the first experiment. The comparison between the two methods can be seen in Figure \ref{fig:training_e4}.

\section{Conclusion}
We have presented a new method for training a feedforward neural network. The core of the method relies on the contraction property of some of the activation function (e.g. sigmoid function) and the geometry underlying the training of FNN. As a result we have obtained a new cost function that need to be minimized during the training process. The main advantage of the new functional resides in a fact that we have less non-linearities introduced by activation functions.

The experiments that we have conducted show that our method results to a faster learning convergence and also to a better recognition rate (classification error). In the first experiment, after 47.750 epochs, the classical method gives a classification error of 3.05 versus our method which renders a classification error of 2.59. This behavior remains valid for the next two experiments, where the classification error for classical method was 2.97 versus 2.00 obtain with the new method, respectively 4.50 classification error obtained with classical method versus 2.79 the classification error obtained with the new method. Even when we have changed the dataset, we did obtain the same improvement by our method that we have seen in the previous experiments. More precisely, the classification error for the classical method was 6.20 versus 4.00 obtained by the new method.
\section*{Compliance with ethical standards} 
{\bf Conflict of interest.} The authors declare that they have no conflict of interest.
\section*{Acknowledgements} 
This work was supported by a grant of Ministery of Research and Innovation,
CNCS - UEFISCDI, project number PN-III-P4-ID-PCE-2016-0165, within PNCDI III.
\bibliographystyle{apalike}
\bibliography{article_bib}   
\end{document}